\documentclass{article} % For LaTeX2e
\usepackage{iclr2016_conference,times}
\usepackage{hyperref}
\usepackage{url}

\usepackage{url}
\usepackage{amsfonts}
\usepackage{amsmath,amsfonts,amssymb,amsthm}
\usepackage{tikz}
\usepackage{pbox}
\usepackage{graphicx}
\usepackage{caption}
\usepackage{subcaption}
\usepackage{graphicx} % leave off demo option in real program
\usepackage{algorithm}
\usepackage{algorithmic}

\title{Incentivizing Exploration In Reinforcement Learning With Deep Predictive Models}
\author{
Bradly C. Stadie \\
Department of Statistics\\
University of California, Berkeley\\
Berkeley, CA 94720 \\
\texttt{bstadie@berkeley.edu} \\
\And
Sergey Levine \hspace{0.1in} Pieter Abbeel\\
EECS Department\\
University of California, Berkeley\\
Berkeley, CA 94720\\
\texttt{\{svlevine,pabbeel\}@cs.berkeley.edu}\\
}

\begin{document}

\maketitle

\begin{abstract}

%%SL4: minor rewording in abstract to remove some words and also remove latex for naive (it makes it a pain to submit abstracts for conference booklet, and generally makes a lot of things a little painful)
Achieving efficient and scalable exploration in complex domains poses a major challenge in reinforcement learning. 
While Bayesian and PAC-MDP approaches to the exploration problem offer strong formal guarantees, they are often impractical in higher dimensions due to their reliance on enumerating the state-action space. Hence, exploration in complex domains is often performed with simple epsilon-greedy methods. In this paper, we consider the challenging Atari games domain, which requires processing raw pixel inputs and delayed rewards. We evaluate several more sophisticated exploration strategies, including Thompson sampling and Boltzman exploration, and propose a new exploration method based on assigning exploration bonuses from a concurrently learned model of the system dynamics. By parameterizing our learned model with a neural network, we are able to develop a scalable and efficient approach to exploration bonuses that can be applied to tasks with complex, high-dimensional state spaces. In the Atari domain, our method provides the most consistent improvement across a range of games that pose a major challenge for prior methods. In addition to raw game-scores, we also develop an AUC-100 metric for the Atari Learning domain to evaluate the impact of exploration on this benchmark.

\end{abstract}

\section{Introduction}
In reinforcement learning (RL), agents acting in unknown environments face the exploration versus exploitation tradeoff. Without adequate exploration, the agent might fail to discover effective control strategies, particularly in complex domains.
Both PAC-MDP algorithms, such as MBIE-EB \cite{pacmdp}, and Bayesian algorithms such as Bayesian Exploration Bonuses (BEB) \cite{beb} have managed this tradeoff by assigning exploration bonuses to novel states. In these methods, the novelty of a state-action pair is derived from the number of times an agent has visited that pair. While these approaches offer strong formal guarantees, their requirement of an enumerable representation of the agent's environment renders them impractical for large-scale tasks. As such, exploration in large RL tasks is still most often performed using simple heuristics, such as the epsilon-greedy strategy \cite{DQN}, which can be inadequate in more complex settings.

In this paper, we evaluate several exploration strategies that can be scaled up to complex tasks with high-dimensional inputs. Our results show that Boltzman exploration and Thompson sampling significantly improve on the na\"{i}ve epsilon-greedy strategy. However, we show that the biggest and most consistent improvement can be achieved by assigning exploration bonuses based on a learned model of the system dynamics with learned representations. To that end, we describe a method that learns a state representation from observations, trains a dynamics model using this representation concurrently with the policy, and uses the misprediction error in this model to asses the novelty of each state. Novel states are expected to disagree more strongly with the model than those states that have been visited frequently in the past, and assigning exploration bonuses based on this disagreement can produce rapid and effective exploration.

%These exploration bonuses are motivated by Bayesian exploration bonuses, in which state-action counts serve to capture the uncertainty in the belief space over a model's transition matrices. Though it is intractable to construct such transition matrices for complex, partially observed tasks with high-dimensional observations such as image pixels, our method captures a similar notion of uncertainty via the misprediction error in the learned dynamics model over the observation space.

Using learned model dynamics to assess a state's novelty presents several challenges. Capturing an adequate representation of the agent's environment for use in dynamics predictions can be accomplished by training a model to predict the next state from the previous ground-truth state-action pair. However, one would not expect pixel intensity values to adequately capture the salient features of a given state-space. To provide a more suitable representation of the system's state space, we propose a method for encoding the state space into lower dimensional domains. To achieve sufficient generality and scalability, we modeled the system's dynamics with a deep neural network. This allows for on-the-fly learning of a model representation that can easily be trained in parallel to learning a policy.

Our main contribution is a scalable and efficient method for assigning exploration bonuses in large RL problems with complex observations, as well as an extensive empirical evaluation of this approach and other simple alternative strategies, such as Boltzman exploration and Thompson sampling. Our approach assigns model-based exploration bonuses from learned representations and dynamics, using only the observations and actions. It can scale to large problems where Bayesian approaches to exploration become impractical, and we show that it achieves significant improvement in learning speed on the task of learning to play Atari games from raw images \cite{ALE}. Our approach achieves state-of-the-art results on a number of games, and achieves particularly large improvements for games on which human players strongly outperform prior methods. Aside from achieving a high final score, our method also achieves substantially faster learning. To evaluate the speed of the learning process, we propose the AUC-100 benchmark to evaluate learning progress on the Atari domain.

%A secondary contribution is the development of the AUC-100 benchmark to evaluate learning progress on the Atari Learning domain. %Say more I'd guess. 

\section{Preliminaries}
\label{prelim}
We consider an infinite-horizon discounted Markov decision process (MDP), defined by the tuple $(\mathcal{S}, \mathcal{A}, \mathcal{P}, \mathcal{R}, \rho_0, \gamma)$, where $\mathcal{S}$ is a finite set of states, $\mathcal{A}$ a finite set of actions, $\mathcal{P} : \mathcal{S} \times A \times \mathcal{S} \to \mathbb{R}$ the transition probability distribution, $\mathcal{R}: \mathcal{S} \to \mathbb{R}$ the reward function, $\rho_0$ an initial state distribution, and $\gamma \in (0,1)$ the discount factor. We are interested in finding a policy $\pi : \mathcal{S} \times \mathcal{A} \to [0,1]$ that maximizes the expected reward over all time. This maximization can be accomplished using a variety of reinforcement learning algorithms.

In this work, we are concerned with online reinforcement learning wherein the algorithm receives a tuple $(s_t,a_t,s_{t+1},r_t)$ at each step. Here, $s_t \in \mathcal{S}$ is the previous state, $a_t \in \mathcal{A}$ is the previous action, $s_{t+1} \in \mathcal{S}$ is the new state, and $r_t$ is the reward collected as a result of this transition. The reinforcement learning algorithm must use this tuple to update its policy and maximize long-term reward and then choose the new action $a_{t+1}$. It is often insufficient to simply choose the best action based on previous experience, since this strategy can quickly fall into a local optimum. Instead, the learning algorithm must perform exploration. Prior work has suggested methods that address the exploration problem by acting with ``optimism under uncertainty.'' If one assumes that the reinforcement learning algorithm will tend to choose the best action, it can be encouraged to visit state-action pairs that it has not frequently seen by augmenting the reward function to deliver a bonus for visiting novel states. This is accomplished with the augmented reward function
\begin{align} 
\mathcal{R}_{\text{Bonus}}(s,a) = \mathcal{R}(s,a) + \beta \mathcal{N}(s,a),
\end{align}
where $\mathcal{N} (s,a) : \mathcal{S} \times \mathcal{A} \to [0,1]$ is a novelty function designed to capture the novelty of a given state-action pair. Prior work has suggested a variety of different novelty functions e.g.,  \cite{pacmdp,beb} based on state visitation frequency.

While such methods offer a number of appealing guarantees, such as near-Bayesian exploration in polynomial time \cite{beb}, they require a concise, often discrete representation of the agent's state-action space to measure state visitation frequencies. In our approach, we will employ function approximation and representation learning to devise an alternative to these requirements. 

%Intuitively, driving an agent towards novel states should require a representation to distinguish the states which the agent has experienced from that which the agent has yet to experience. However, using the simple tabular approach for measuring novelty via visitation frequency becomes intractable as the dimensionality of the state-action space increases. Furthermore, it fails to capture similarity between state-action pairs that are distinct but represent semantically similar situations. In our approach, we will employ function approximation and representation learning to address these challenges.

\section{Model Learning For Exploration Bonuses} 

We would like to encourage agent exploration by giving the agent exploration bonuses for visiting novel states. Identifying states as novel requires we supply some representation of the agent's state space, as well as a mechanism to use this representation to assess novelty. Unsupervised learning methods offer one promising avenue for acquiring a concise representation of the state with a good similarity metric. This can be accomplished using dimensionality reduction, clustering, or graph-based techniques \cite{ML, graphlap}. In our work, we draw on recent developments in representation learning with neural networks, as discussed in the following section. However, even with a good learned state representation, maintaining a table of visitation frequencies becomes impractical for complex tasks. Instead, we learn a model of the task dynamics that can be used to assess the novelty of a new state.

Formally, let $\sigma(s)$ denote the encoding of the state $s$, and let $\mathcal{M}_{\phi} : \mathcal{ \sigma(S) } \times \mathcal{A} \to \mathcal{ \sigma(S) }$ be a dynamics predictor parameterized by $\phi$. $\mathcal{M}_\phi$ takes an encoded version of a state $s$ at time $t$ and the agent's action at time $t$ and attempts to predict an encoded version of the agent's state at time $t+1$. The parameterization of $\mathcal{M}$ is discussed further in the next section.

For each state transition $(s_t,a_t,s_{t+1})$, we can attempt to predict $\sigma(s_{t+1})$ from $(\sigma(s_t),a_t)$ using our predictive model $\mathcal{M}_\phi$. This prediction will have some error
\begin{align} 
e(s_t, a_t) =  \| \sigma(s_{t+1}) - \mathcal{M}_{\phi} (\sigma(s_t), a_t)   \|_2^2.
\end{align}
Let $\overline{e_T}$, the normalized prediction error at time $T$, be given by $\overline{e_T} := \frac{e_T} {\max_{t \leq T}  \{e_t \}}$. We can assign a novelty function to $(s_t, a_t)$ via
\begin{align} 
\mathcal{N}(s_t, a_t) &=    \frac{ \bar{e}_t (s_t, a_t) }{t \ast C} 
\end{align}
where $C >0$ is a decay constant. We can now realize our augmented reward function as
\begin{align}
\mathcal{R}_{Bonus} (s,a) = \mathcal{R}(s,a) + \beta  \left( \frac { \bar{e}_t (s_t, a_t) }{t \ast C} \right)\label{eqn:bonus}
\end{align}

This approach is motivated by the idea that, as our ability to model the dynamics of a particular state-action pair improves, we have come to understand the state better and hence its novelty is lower. When we don't understand the state-action pair well enough to make accurate predictions, we assume that more knowledge about that particular area of the model dynamics is needed and hence a higher novelty measure is assigned. 

Using learned model dynamics to assign novelty functions allows us to address the exploration versus exploitation problem in a non-greedy way. With an appropriate representation $\sigma(s_t)$, even when we encounter a new state-action pair $(s_t, a_t)$, we expect $\mathcal{M}_\phi (\sigma(s_t), a_t)$ to be accurate so long as enough similar state-action pairs have been encountered.

\footnotesize
\begin{algorithm}[tb]
\caption{Reinforcement learning with model prediction exploration bonuses}
\label{alg:summary}
\begin{algorithmic}[1]
\STATE Initialize $\max_e = 1$, $\text{EpochLength}$, $\beta$, $C$
\FOR{iteration $t$ in $T$}
\STATE Observe $(s_t, a_t, s_{t+1}, \mathcal{R}(s_t, a_t))$
\STATE Encode the observations to obtain $\sigma(s_t)$ and $\sigma(s_{t+1})$
\STATE Compute $e(s_t, a_t) = \| \sigma(s_{t+1}) - \mathcal{M}_{\phi} (\sigma(s_t), a_t) \|_2^2$ and $\bar{e}(s_t,a_t) = \frac{e(s_t, a_t)}{\max_e}$.
\STATE Compute $\mathcal{R}_{Bonus} (s_t,a_t) = \mathcal{R}(s,a) + \beta  \left(  \frac{ \bar{e}_t (s_t, a_t) }{t \ast C}  \right)$
\IF{ $e(s_t, a_t) > \max_e$}
\STATE $\max_e = e(s_t, a_t)$
\ENDIF
\STATE Store $(s_t, a_t, \mathcal{R}_{bonus})$ in a memory bank $\Omega$. 
\STATE Pass $\Omega$  to the reinforcement learning algorithm to update $\pi$. 
\IF{ $ t \mod \text{EpochLength} == 0$}
\STATE Use $\Omega$ to update $\mathcal{M}$.
\STATE Optionally, update $\sigma$.
\ENDIF
\ENDFOR
\STATE {\bf return} optimized policy $\pi$
\end{algorithmic}
\end{algorithm}
\normalsize

Our model-based exploration bonuses can be incorporated into any online reinforcement learning algorithm that updates the policy based on state, action, reward tuples of the form $(s_t, a_t, s_{t+1}, r_t)$, such as Q-learning or actor-critic algorithms. Our method is summarized in Algorithm~\ref{alg:summary}. At each step, we receive a tuple $(s_t, a_t, s_{t+1}, \mathcal{R}(s_t, a_t))$ and compute the Euclidean distance between the encoded state $\sigma(s_{t+1})$ to the prediction made by our model $\mathcal{M}_\phi (\sigma(s_t),a_t)$. This is used to compute the exploration-augmented reward $\mathcal{R}_{Bonus}$ using Equation (\ref{eqn:bonus}). The tuples $(s_t, a_t, s_{t+1}, \mathcal{R}_{\text{Bonus}})$ are stored in a memory bank $\Omega$ at the end of every step.  Every step, the policy is updated. \footnote{In our implementation, the memory bank $\Omega$ is used to retrain the RL algorithm via experience replay once per epoch (50000 steps). Hence, 49999 of these policy updates will simply do nothing.} Once per epoch, corresponding to 50000 observations in our implementation, the dynamics model $\mathcal{M}_\phi$ is updated to improve its accuracy. If desired, the representation encoder $\sigma$ can also be updated at this time. We found that retraining $\sigma$ once every 5 epochs to be sufficient.
%%SL4: are the above details specific to deep Q learning? if so, should they be moved somewhere else? My impression was that sec 3 was meant to stay "generic" and general to any choice of learning algorithm...

%%SL4: commenting this out, since it's repeated twice already
%The particular form of the model and encoder updates depends on the parameterization of $\mathcal{M}$ and $\sigma$, which we discuss in the following section.

This approach is modular and compatible with any representation of $\sigma$ and $\mathcal{M}$, as well as any reinforcement learning method that updates its policy based on a continuous stream of observation, action, reward tuples. Incorporating exploration bonuses does make the reinforcement learning task nonstationary, though we did not find this to be a major issue in practice, as shown in our experimental evaluation. In the following section, we discuss the particular choice for $\sigma$ and $\mathcal{M}$ that we use for learning policies for playing Atari games from raw images.

\section{Deep Learning Architectures} 

Though the dynamics model $\mathcal{M}_\phi$ and the encoder $\sigma$ from the previous section can be parametrized by any appropriate method, we found that using deep neural networks for both achieved good empirical results on the Atari games benchmark. In this section, we discuss the particular networks used in our implementation.

\subsection{Autoencoders} 

The most direct way of learning a dynamics model is to directly predict the state at the next time step, which in the Atari games benchmark corresponds to the next frame's pixel intensity values. However, directly predicting these pixel intensity values is unsatisfactory, since we do not expect pixel intensity to capture the salient features of the environment in a robust way. In our experiments, a dynamics model trained to predict raw frames exhibited extremely poor behavior, assigning exploration bonuses in near equality at most time steps, as discussed in our experimental results section.

To overcome these difficulties, we seek a function $\sigma$ which encodes a lower dimensional representation of the state $s$.  For the task of representing Atari frames, we found that an autoencoder could be used to successfully obtain an encoding function $\sigma$ and achieve dimensionality reduction and feature extraction \cite{hinton}. 
\begin{figure}[!htbp]
\begin{center}
\includegraphics[width=0.62\textwidth]{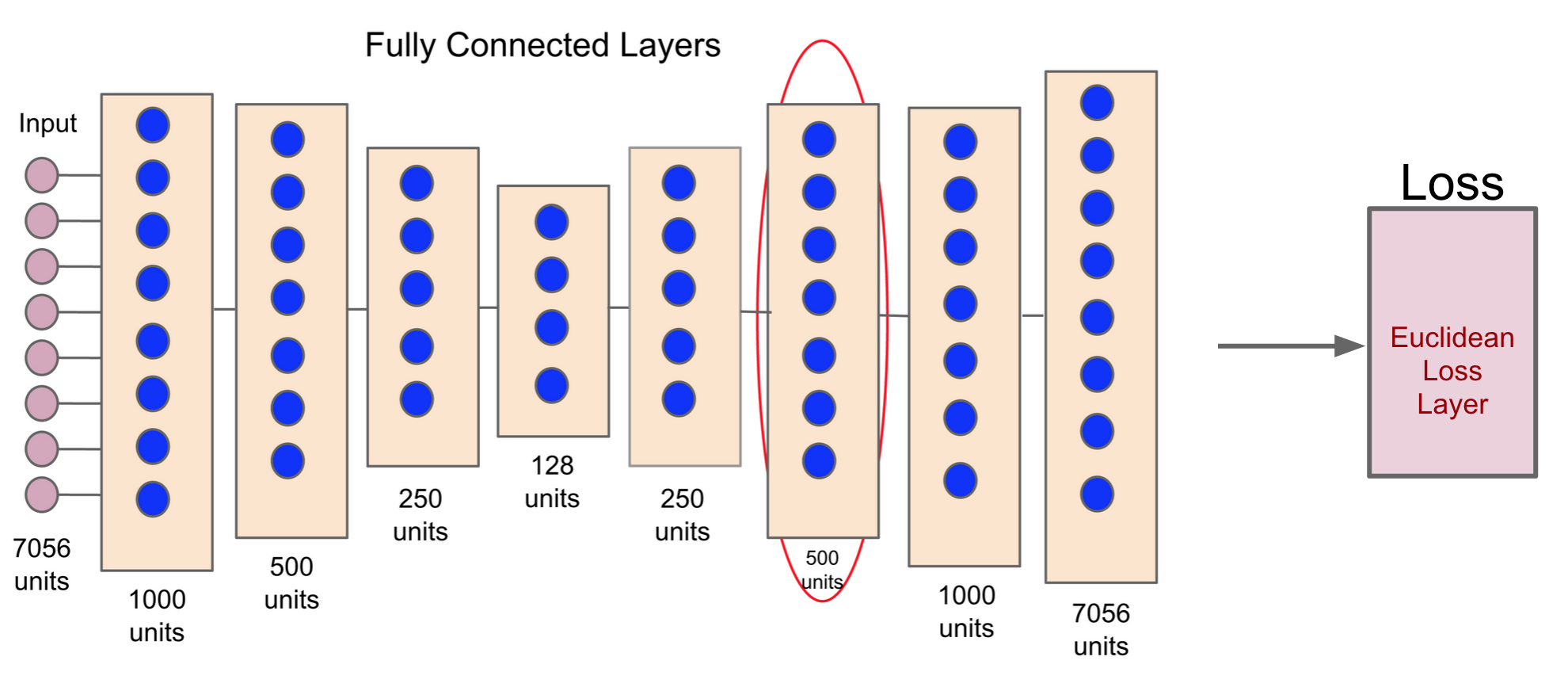} \hspace{5mm}
\includegraphics[width=0.33\textwidth]{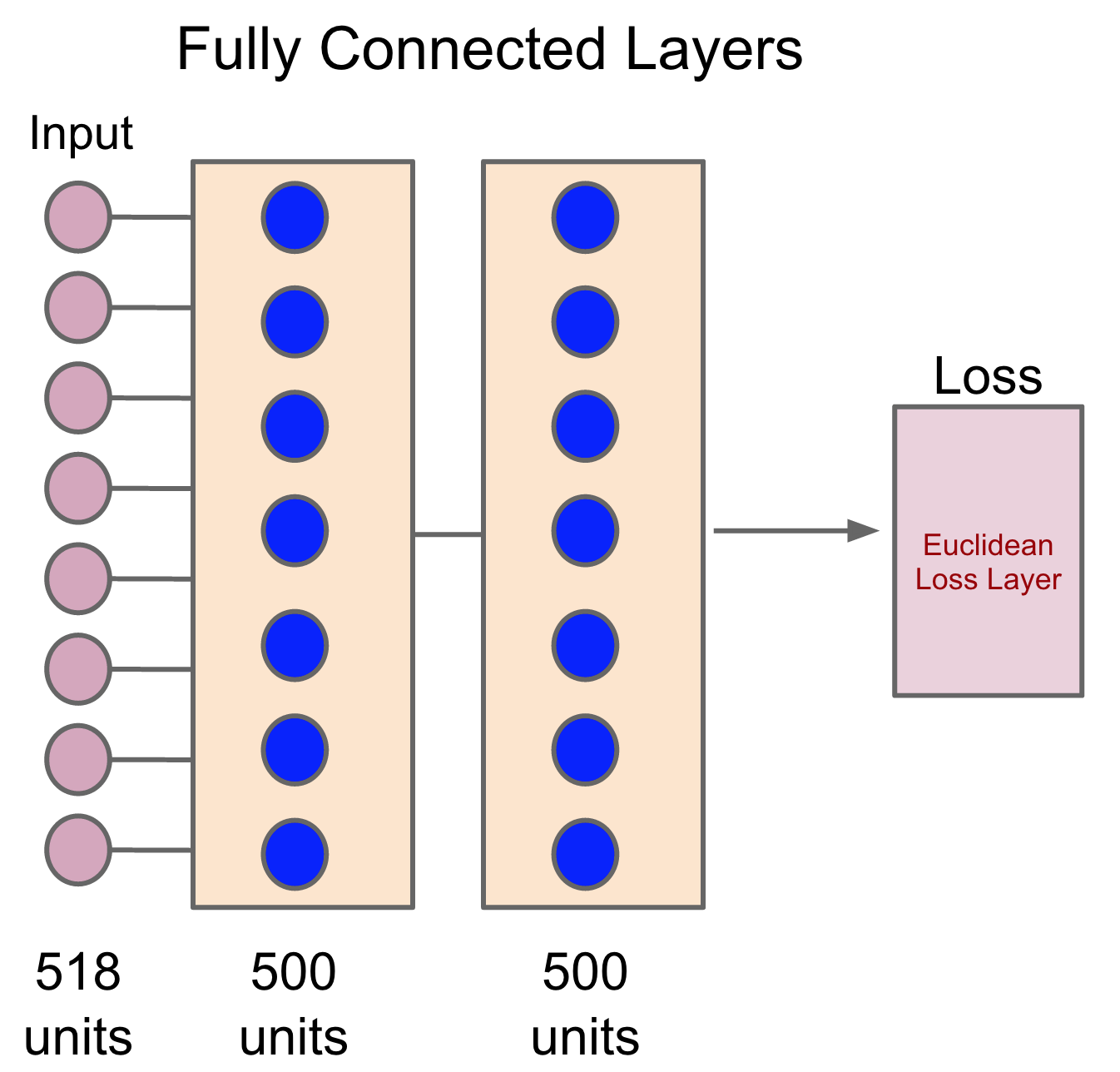} 
\end{center}
\caption{Left: Autoencoder used on input space. The circle denotes the hidden layer that was extracted and utilized as input for dynamics learning. Right: Model learning architecture.}
\end{figure}
Our autoencoder has 8 hidden layers, followed by a Euclidean loss layer, which computes the distance between the output features and the original input image. The hidden layers are reduced in dimension until maximal compression occurs with 128 units. After this, the activations are decoded by passing through hidden layers with increasingly large size. We train the network on a set of 250,000 images and test on a further set of 25,000 images. We compared two separate methodologies for capturing these images. 
\begin{enumerate} 
\item \textbf{Static AE:} A random agent plays for enough time to collect the required images. The auto-encoder $\sigma$ is trained offline before the policy learning algorithm begins. 
\item \textbf{Dynamic AE:} Initialize with an epsilon-greedy strategy and collect images and actions while the agent acts under the policy learning algorithm. After 5 epochs, train the auto encoder from this data. Continue to collect data and periodically retrain the auto encoder in parallel with the policy training algorithm. 

\end{enumerate}
We found that the reconstructed input achieves a small but non-trivial residual on the test set regardless of which auto encoder training technique is utilized, suggesting that in both cases it learns underlying features of the state space while avoiding overfitting.

To obtain a lower dimensional representation of the agent's state space, a snapshot of the network's first six layers is saved. The sixth layer's output (circled in figure one) is then utilized as an encoding for the original state space. That is, we construct an encoding $\sigma(s_t)$ by running $s_t$ through the first six hidden layers of our autoencoder and then taking the sixth layers output to be $\sigma(s_t)$. In practice, we found that using the sixth layer's output (rather than the bottleneck at the fifth layer) obtained the best model learning results. See the appendix for further discussion on this result. 

\subsection{Model Learning Architecture}
Equipped with an encoding $\sigma$, we can now consider the task of predicting model dynamics. For this task, a much simpler two layer neural network $\mathcal{M}_\phi$ suffices. $\mathcal{M}_\phi$ takes as input the encoded version of a state $s_t$ at time $t$ along with the agent's action $a_t$ and seeks to predict the encoded next frame $\sigma(s_{t+1})$. Loss is computed via a Euclidean loss layer regressing on the ground truth $\sigma(s_{t+1})$. We find that this model initially learns a representation close to the identity function and consequently the loss residual is similar for most state-action pairs. However, after approximately 5 epochs, it begins to learn more complex dynamics and consequently better identify novel states. We evaluate the quality of the learned model in the appendix. 

%\begin{figure}[!htbp]
%\begin{center}
%\includegraphics[width=0.33\textwidth]{model_dynamics} 
%\end{center}
%\caption{Architecture for learning model dynamics. Since the state space is already encoded, a relatively simple architecture is able to robustly capture the system's dynamics.}
%\end{figure}

\section{Related Work} 

Exploration is an intensely studied area of reinforcement learning. Many of the pioneering algorithms in this area, such as $R-Max$ \cite{rmax} and $E^3$ \cite{req3}, achieve efficient exploration that scales polynomially with the number of parameters in the agent's state space (see also \cite{req1,  req4}). However, as the size of state spaces increases, these methods quickly become intractable. A number of prior methods also examine various techniques for using models and prediction to incentivize exploration \cite{variance,progress,valuefunctionapprox, lastmin1}. However, such methods typically operate directly on the transition matrix of a discrete MDP, and do not provide for a straightforward extension to very large or continuous spaces, where function approximation is required. A number of prior methods have also been proposed to incorporate domain-specific factors to improve exploration. Doshi-Velez et al. \cite{bayesianpolicypriors} proposed incorporating priors into policy optimization, while Lang et al. \cite{relationaldomains} developed a method specific to relational domains. Finally,  Schmidhuber et al. have developed a curiosity driven approach to exploration which uses model predictors to aid in control \cite{schmidhuber}. 

Several exploration techniques have been proposed that can extend more readily to large state spaces. Among these, methods such as C-PACE \cite{pacoptimal} and metric-$E^3$ \cite{req2} require a good metric on the state space that satisfies the assumptions of the algorithm. The corresponding representation learning issue has some parallels to the representation problem that we address by using an autoecoder, but it is unclear how the appropriate metric for the prior methods can be acquired automatically on tasks with raw sensory input, such as the Atari games in our experimental evaluation. Methods based on Monte-Carlo tree search can also scale gracefully to complex domains \cite{samplebasedsearch}, and indeed previous work has applied such techniques to the task of playing Atari games from screen images \cite{Guo}. However, this approach is computationally very intensive, and requires access to a generative model of the system in order to perform the tree search, which is not always available in online reinforcement learning. On the other hand, our method readily integrates into any online reinforcement learning algorithm.

Finally, several recent papers have focused on driving the Q value higher. In \cite{Gal}, the authors use network dropout to perform Thompson sampling. In Boltzman exploration, a positive probability is assigned to any possible action according to its expected utility and according to a temperature parameter \cite{Boltzman}. Both of these methods focus on controlling Q values rather than model-based exploration. A comparison to both is provided in the next section.

\section{Experimental Results} 

We evaluate our approach on 14 games from the Arcade Learning Environment \cite{ALE}. The task consists of choosing actions in an Atari emulator based on raw images of the screen. Previous work has tackled this task using Q-learning with epsilon-greedy exploration \cite{DQN}, as well as Monte Carlo tree search \cite{Guo} and policy gradient methods \cite{TRPO}. We use Deep Q Networks (DQN) \cite{DQN} as the reinforcement learning algorithm within our method, and compare its performance to the same DQN method using only epsilon-greedy exploration, Boltzman exploration, and a Thompson sampling approach.

The results for 14 games in the Arcade Learning Environment are presented in Table 1. We chose those games that were particularly challenging for prior methods and ones where human experts outperform prior learning methods. We evaluated two versions of our approach; using either an autoencoder trained in advance by running epsilon-greedy Q-learning to collect data (denoted as ``Static AE''), or using an autoencoder trained concurrently with the model and policy on the same image data (denoted as ``Dynamic AE''). Table 1 also shows results from the DQN implementation reported in previous work, along with human expert performance on each game \cite{DQN}. Note that our DQN implementation did not attain the same score on all of the games as prior work due to a shorter running time. Since we are primarily concerned with the rate of learning and not the final results, we do not consider this a deficiency. To directly evaluate the benefit of including exploration bonuses, we compare the performance of our approach primarily to our own DQN implementation, with the prior scores provided for reference.

In addition to raw-game scores, and learning curves, we also analyze our results on a new benchmark we have named Area Under Curve 100 (AUC-100). For each game, this benchmark computes the area under the game-score learning curve (using the trapezoid rule to approximate the integral). This area is then normalized by 100 times the score maximum game score achieved in \cite{DQN}, which represents 100 epochs of play at the best-known levels. This metric more effectively captures improvements to the game's learning rate and does not require running the games for 1000 epochs as in \cite{DQN}. For this reason, we suggest it as an alternative metric to raw game-score.

\begin{figure}[!htbp]
\label{curves}
\begin{center}$
\begin{array}{cccc}
\includegraphics[width=0.29\textwidth]{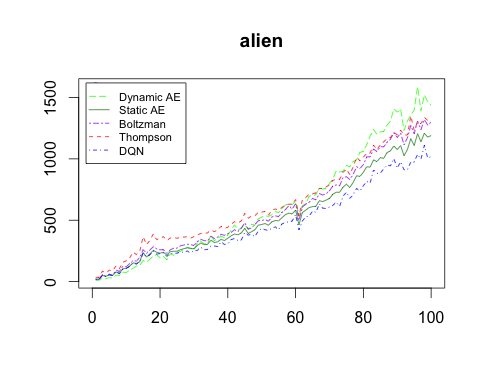} &
\includegraphics[width=0.29\textwidth]{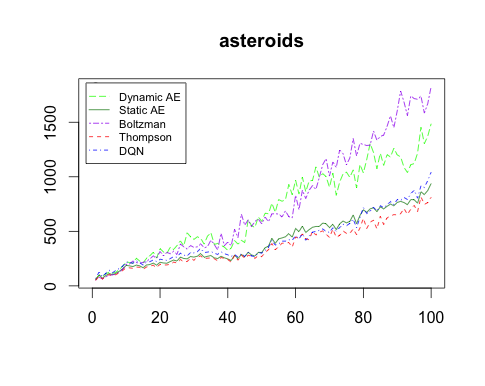} &
\includegraphics[width=0.29\textwidth]{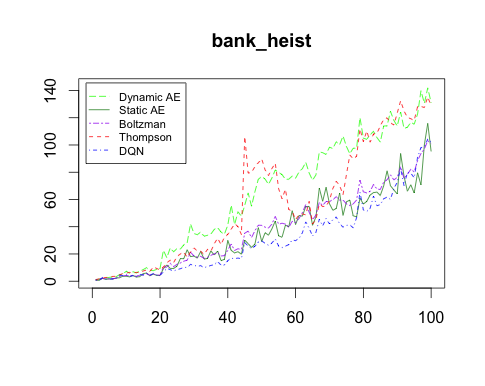} &
\includegraphics[width=0.29\textwidth]{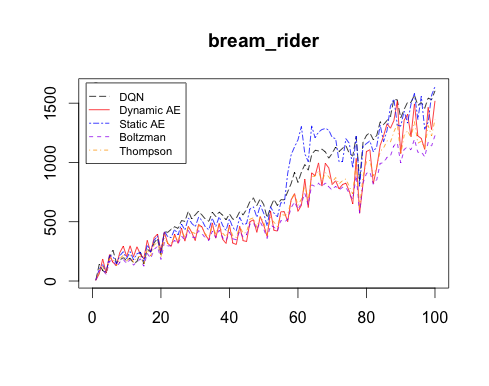} \\
\includegraphics[width=0.29\textwidth]{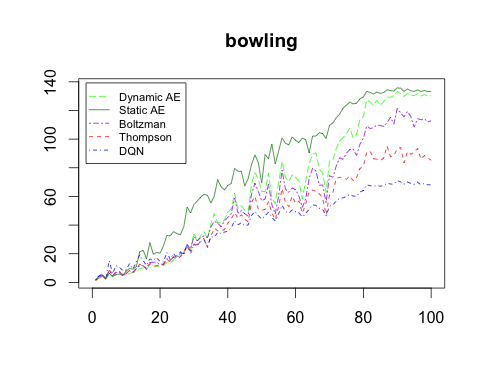} &
\includegraphics[width=0.29\textwidth]{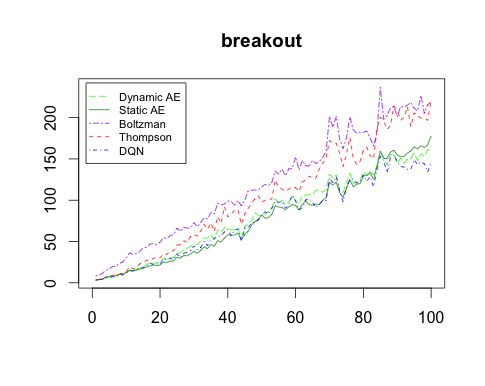} &
\includegraphics[width=0.29\textwidth]{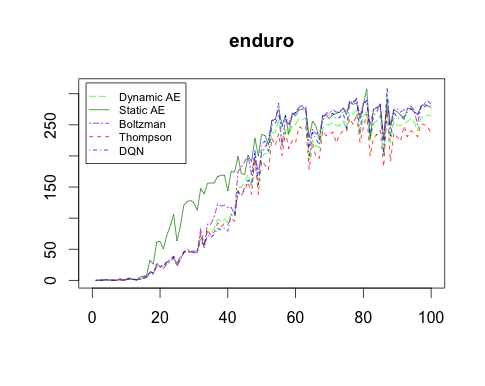} &
\includegraphics[width=0.29\textwidth]{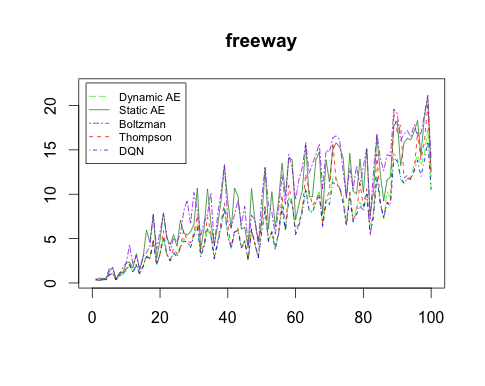} \\
\includegraphics[width=0.29\textwidth]{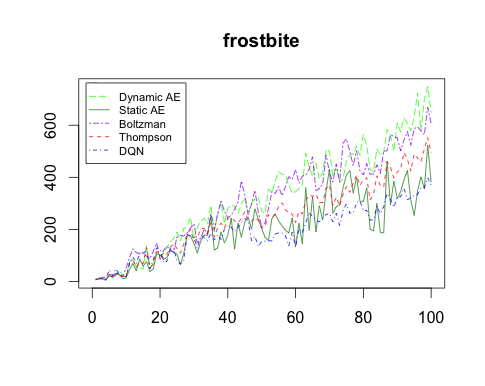} &
\includegraphics[width=0.29\textwidth]{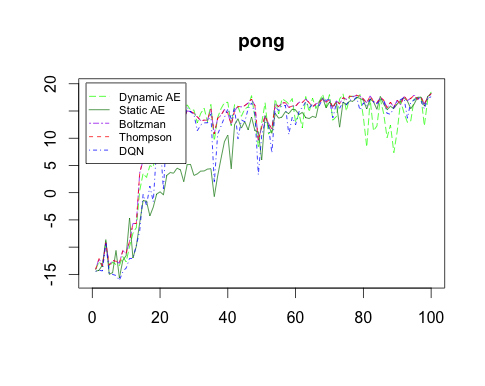} &
\includegraphics[width=0.29\textwidth]{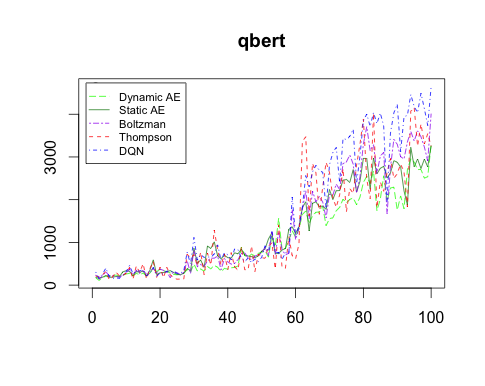} &
\includegraphics[width=0.29\textwidth]{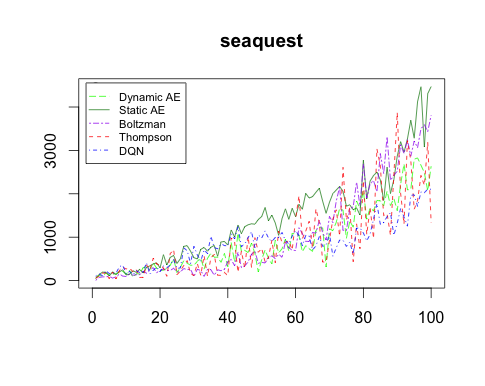} \\
\includegraphics[width=0.29\textwidth]{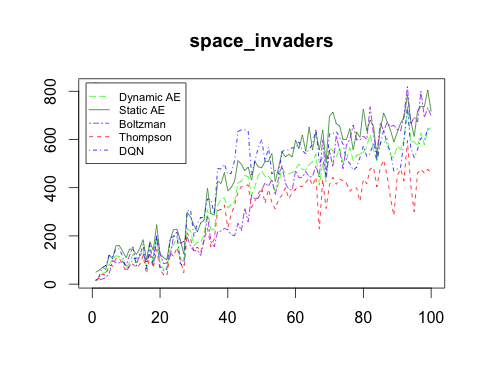} &
\includegraphics[width=0.24\textwidth]{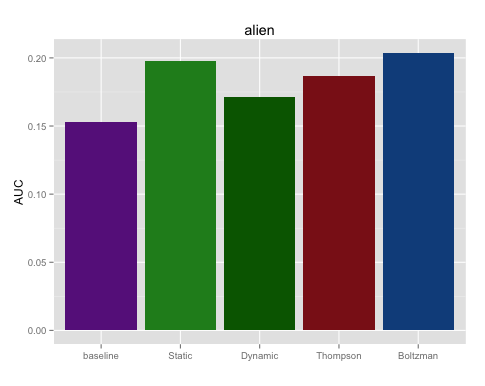} &
\includegraphics[width=0.24\textwidth]{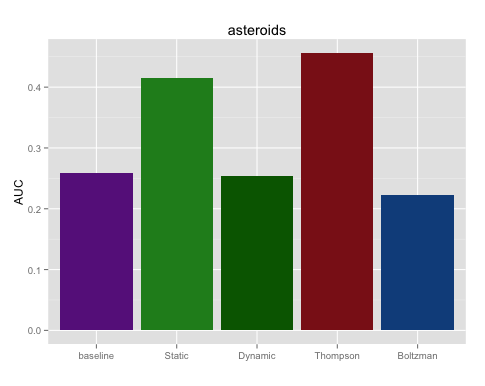} &
\includegraphics[width=0.24\textwidth]{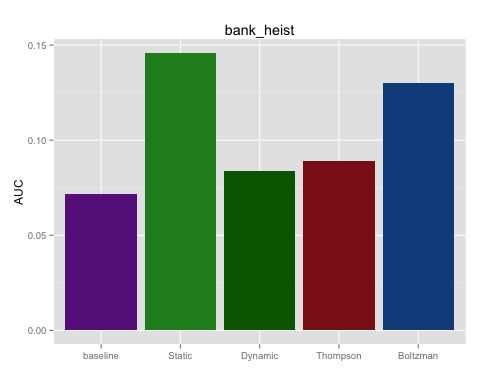} \\
\includegraphics[width=0.24\textwidth]{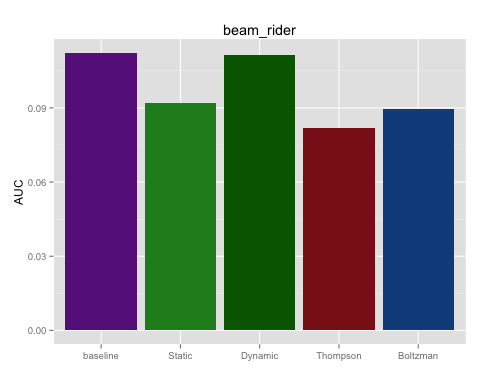} &
\includegraphics[width=0.24\textwidth]{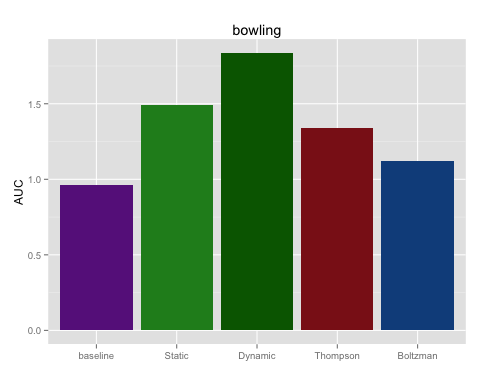} &
\includegraphics[width=0.24\textwidth]{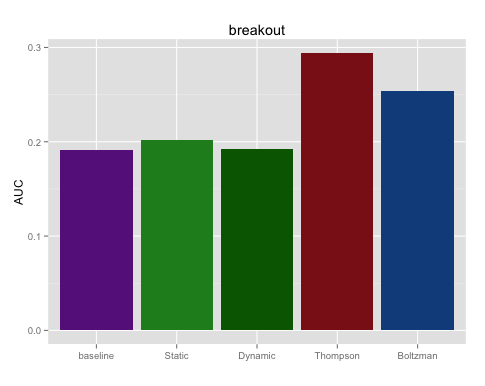} &
\includegraphics[width=0.24\textwidth]{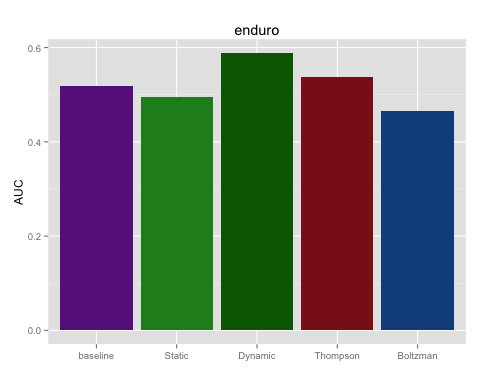} \\
\includegraphics[width=0.24\textwidth]{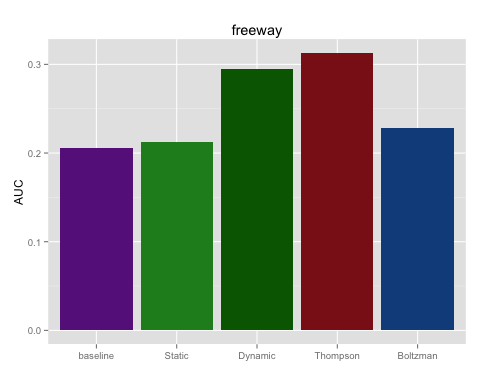} &
\includegraphics[width=0.24\textwidth]{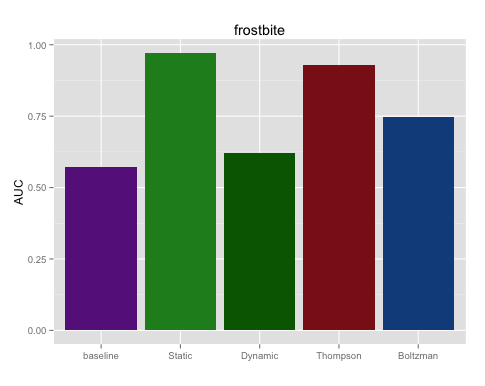} &
\includegraphics[width=0.24\textwidth]{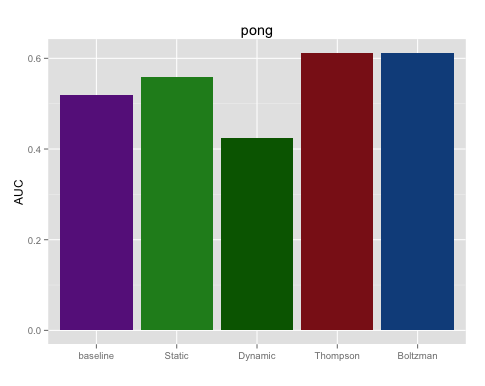} &
\includegraphics[width=0.24\textwidth]{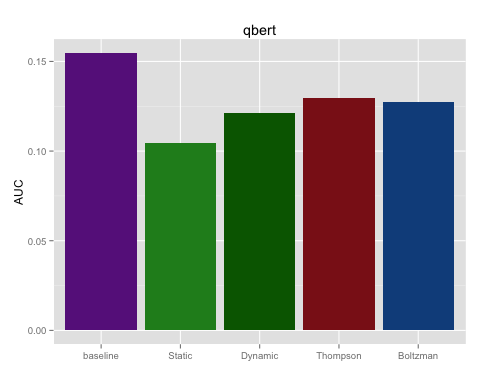} \\
\includegraphics[width=0.24\textwidth]{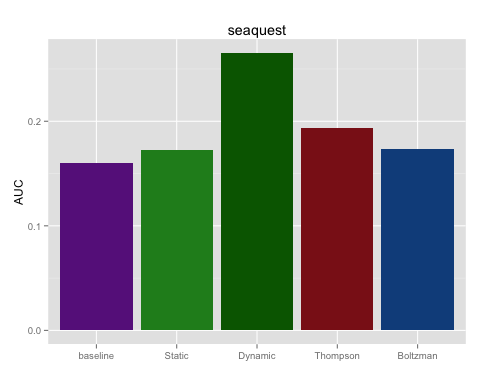} &
\includegraphics[width=0.24\textwidth]{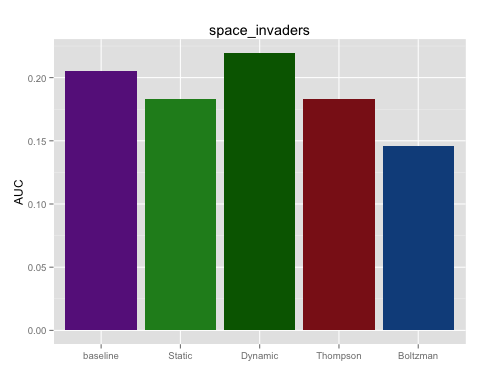} &
\end{array}$
\end{center}
\caption{Full learning curves and AUC-100 scores for all Atari games. We present the raw AUC-100 scores in the appendix.}
\end{figure}

%Exploration bonuses improved the final results for nearly every game, except for $\text{Q}^{\ast}$bert. Exploration bonuses achieved the biggest improvement on Asteroids (148 \% improvement), Bowling (136\%), Frostbite (148\%), Freeway (252 \%), and Seaquest (133\%). We discuss some of these games below.

\paragraph{Bowling} The policy without exploration tended to fixate on a set pattern of nocking down six pins per frame. When bonuses were added, the dynamics learner quickly became adept at predicting this outcome and was thus encouraged to explore other release points.

\paragraph{Frostbite} This game's dynamics changed substantially via the addition of extra platforms as the player progressed. As the dynamics of these more complex systems was not well understood, the system was encouraged to visit them often (which required making further progress in the game). 

\paragraph{Seaquest} A submarine must surface for air between bouts of fighting sharks. However, if the player resurfaces too soon they will suffer a penalty with effects on the game's dynamics. Since these effects  are poorly understood by the model learning algorithm, resurfacing receives a high exploration bonus and hence the agent eventually learns to successfully resurface at the correct time.

\paragraph{$\text{Q}^{\ast}$bert} Exploration bonuses resulted in a lower score. In $\text{Q}^{\ast}$bert, the background changes color after level one. The dynamics predictor is unable to quickly adapt to such a dramatic change in the environment and consequently, exploration bonuses are assigned in near equality to almost every state that is visited. This negatively impacts the final policy.

%On the game Seaquest, we also evaluated the performance of the na\"{i}ve identity encoding of the state, where the model attempts to directly predict the pixels in the next image. Because nearly every image is different in a subtle way, this type of encoding performed quite poorly, achieving a final score of 252. This illustrates the importance of choosing an appropriate state representation for prediction.

Learning curves for each of the games are shown in Figure (3). Note that both of the exploration bonus algorithms learn significantly faster than epsilon-greedy Q-learning, and often continue learning even after the epsilon-greedy strategy converges. All games had the inputs normalized according to \cite{DQN} and were run for 100 epochs (where one epoch is 50,000 time steps). Between each epoch, the policy was updated and then the new policy underwent 10,000 time steps of testing. The results represent the average testing score across three trials after 100 epoch each.
\begin{table}[ht!]
\small
\begin{center}
\begin{tabular}{l|lllll||ll}
\multicolumn{1}{c}{\bf Game}  &\multicolumn{1}{c}{ \pbox{20cm}{\bf DQN \\ 100 epochs}} &\multicolumn{1}{c}{\pbox{95cm}{\bf Exploration \\ Static AE \\ 100 epochs}} &\multicolumn{1}{c}{\pbox{25cm}{\bf Exploration \\ Dynamic AE \\ 100 epochs}} & \multicolumn{1}{c}{\bf \pbox{20cm}{Boltzman \\ Exploration \\ 100 epochs} }& \multicolumn{1}{c}{\bf \pbox{20cm}{Thompson \\ Sampling \\ 100 epochs} }& \multicolumn{1}{c}{\bf \pbox{20cm}{DQN \cite{DQN} \\ 1000 epochs} }  &\multicolumn{1}{c}{\pbox{95cm}{\bf Human \\ Expert \cite{DQN}}} 
\\ \hline\\
Alien     &   1018 & \textbf{1436} & 1190  &1301 &1322 & 3069 &6875 \\
Asteroids         & 1043 &\textbf{1486} & 939  & 1287 &812 &1629 &13157 \\
Bank Heist      &102 & \textbf{131} & 95  & 101 &129 &   429.7 &734.4   \\
Beam Rider      & 1604 & 1520 & \textbf{1640}   & 1228 &1361 & 6846 &5775  \\
Bowling      & 68.1 & 130 & \textbf{133}  & 113 &85.2 & 42.4 &154.8   \\
Breakout     & 146 & 162 & 178  & 219 & \textbf{222} & 401.2  &31.8\\
Enduro        & 281 & 264 & 277 & \textbf{284} &236 & 301.8 & 309.6  \\
Freeway     & 10.5 & 10.5 & 12.5  & \textbf{13.9} &12.0 & 30.3  & 29.6   \\
Frostbite      &369 & \textbf{649} & 380 & 605 &494 & 328.3   &4335  \\
Montezuma & 0.0 & 0.0 & 0.0 & 0 &0 & 0.0 & 4367  \\
Pong           & 17.6 & \textbf{18.5} & 18.2 & 18.2 &18.2 & 18.9 &9.3  \\
$\text{Q}^\ast$bert       & \textbf{4649} & 3291 &3263  & 4014 &3251 & 10596 &13455  \\
Seaquest         & 2106 & 2636 & \textbf{4472} & 3808 &1337 & 5286 &20182  \\
Space Invaders   & 634 & 649 &  \textbf{716}   & 697 &459 & 1976 &1652   \\
\end{tabular}
\end{center}
\caption{\footnotesize A comparison of maximum scores achieved by different methods. Static AE trains the state-space auto encoder on 250000 raw game frames prior to policy optimization (raw frames are taken from random agent play). Dynamic AE retrains the auto encoder after each epoch, using the last 250000 images as a training set.  Note that exploration bonuses help us to achieve state of the art results on Bowling and Frostbite. Each of these games provides a significant exploration challenge. Bolded numbers indicate the best-performing score among our experiments. Note that this score is sometimes lower than the score reported for DQN in prior work as our implementation only one-tenth as long as in \cite{DQN}.} 
\label{Full Table}
\end{table}
\normalsize

The results show that more nuanced exploration strategies generally improve on the naive epsilon greedy approach, with the Boltzman and Thompson sampling methods achieving the best results on three of the games. However, exploration bonuses achieve the fastest learning and the best results most consistently, outperforming the other three methods on 7 of the 14 games in terms of AUC-100. 

\section{Conclusion} 

In this paper, we evaluated several scalable and efficient exploration algorithms for reinforcement learning in tasks with complex, high-dimensional observations. Our results show that a new method based on assigning exploration bonuses most consistently achieves the largest improvement on a range of challenging Atari games, particularly those on which human players outperform prior learning methods. Our exploration method learns a model of the dynamics concurrently with the policy. This model predicts a learned representation of the state, and a function of this prediction error is added to the reward as an exploration bonus to encourage the policy to visit states with high novelty.

One of the limitations of our approach is that the misprediction error metric assumes that any misprediction in the state is caused by inaccuracies in the model. While this is true in determinstic environments, stochastic dynamics violate this assumption. An extension of our approach to stochastic systems requires a more nuanced treatment of the distinction between \emph{stochastic} dynamics and \emph{uncertain} dynamics, which we hope to explore in future work. Another intriguing direction for future work is to examine how the learned dynamics model can be incorporated into the policy learning process, beyond just providing exploration bonuses. This could in principle enable substantially faster learning than purely model-free approaches.

\newpage

%\scriptsize

\footnotesize
\section{Appendix} 

\subsection{On auto encoder layer selection} 
Recall that we trained an auto-encoder to encode the game's state space. We then trained a predictive model on the next auto-encoded frame rather than directly training on the pixel intensity values of the next frame. To obtain the encoded space, we ran each state through an eight layer auto-encoder for training and then utilized the auto-encoder's sixth layer as an encoded state space. We chose to use the sixth layer rather than the bottleneck fourth layer because we found that, over 20 iterations of Seaquest at 100 epochs per iteration, using this layer for encoding delivered measurably better performance than using the bottleneck layer. The results of that experiment are presented below. 
\begin{figure}[!htbp]
\begin{center}
\includegraphics[width=0.41\textwidth]{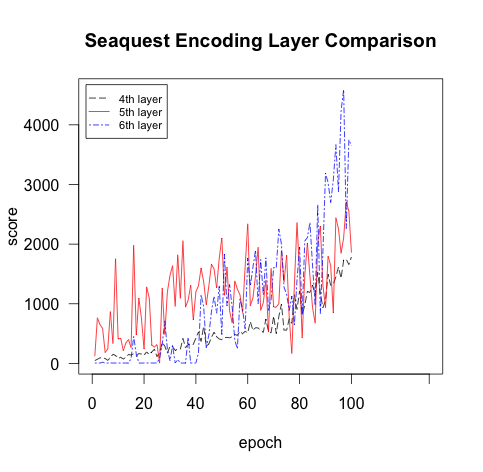} 
\end{center}
\caption{Game score averaged over 20 Seaquest iterations with various choices for the state-space encoding layer. Notice that choosing the sixth layer to encode the state space significantly outperformed the bottleneck layer.}
\end{figure}

\subsection{On the quality of the learned model dynamics} 
Evaluating the quality of the learned dynamics model is somewhat difficult because the system is rewarded achieving higher error rates. A dynamics model that converges quickly is not useful for exploration bonuses. Nevertheless, when we plot the mean of the normalized residuals across all games and all trials used in our experiments, we see that the errors of the learned dynamics models continually decrease over time. The mean normalized residual after 100 epochs is approximately half of the maximal mean achieved. This suggests that each dynamics model was able to correctly learn properties of underlying dynamics for its given game. 
\begin{figure}[!htbp]
\begin{center}
\includegraphics[width=0.41\textwidth]{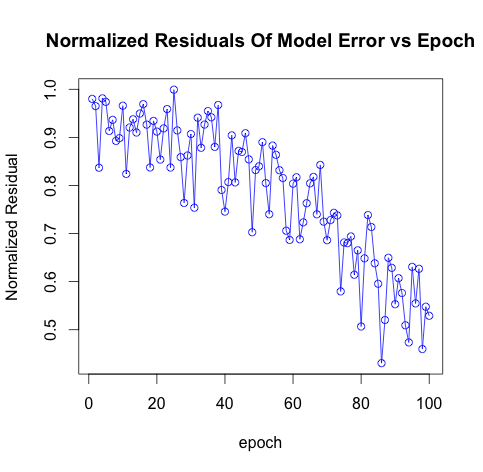} 
\end{center}
\caption{Normalized dynamics model prediction residual across all trials of all games. Note that the dynamics model is retrained from scratch for each trial.}
\end{figure}

\subsection{Raw AUC-100 scores} 

\begin{table}[ht!]
\begin{center}
\begin{tabular}{l|lll||llll}
\multicolumn{1}{c}{\bf Game}  &\multicolumn{1}{c}{ \pbox{20cm}{\bf DQN}} &\multicolumn{1}{c}{\pbox{95cm}{\bf Exploration \\ Static AE}} &\multicolumn{1}{c}{\pbox{25cm}{\bf Exploration \\ Dynamic AE}} & \multicolumn{1}{c}{\bf \pbox{20cm}{Boltzman \\ Exploration} }& \multicolumn{1}{c}{\bf \pbox{20cm}{Thompson \\ Sampling} }
\\ \hline\\
Alien     &   0.153 & 0.198 & 0.171 &  0.187 & 0.204  \\
Asteroids         &  0.259 & 0.415  & 0.254 & 0.456 & 0.223  \\
Bank Heist      &0.0715 & 0.1459 & 0.089  & 0.089 &0.1303  \\
Beam Rider     & 0.1122 &  0.0919 & 0.1112 & 0.0817   & 0.0897   \\
Bowling      &  0.964 & 1.493 & 1.836  & 1.338 & 1.122    \\
Breakout     &  0.191& 0.202 & 0.192  & 0.294 & 0.254 \\
Enduro        & 0.518 & 0.495 & 0.589 & 0.538 & 0.466 \\
Freeway     & 0.206 & 0.213 & 0.295  & 0.313 &0.228   \\
Frostbite      &0.573 & 0.971 & 0.622 & 0.928 &0.746  \\
Montezuma & 0.0 & 0.0 & 0.0 & 0 &0  \\
Pong           & 0.52 & 0.56 & 0.424 & 0.612 & 0.612   \\
$\text{Q}^\ast$bert       & 0.155 & 0.104 & 0.121  & 0.13 & 0.127  \\
Seaquest         & 0.16 & 0.172 &  0.265 & 0.194 & 0.174  \\
Space Invaders   & 0.205 & 0.183 &  0.219   & 0.183 & 0.146 \\
\end{tabular}
\end{center}
\caption{AUC-100 is computed by comparing the area under the game-score learning curve for 100 epochs of play to the area under of the rectangle with dimensions 100 by the maximum DQN score the game achieved in \cite{DQN}. The integral is approximated with the trapezoid rule. This more holistically captures the games learning rate and does not require running the games for 1000 epochs as in \cite{DQN}. For this reason, we suggest it as an alternative metric to raw game-score.}
\label{Full Table}
\end{table}

%\subsection{Full Results} 
%Below, we present the table and learning curves over all 14 games in the Atari learning domain. We see that our method preforms significantly better in games where exploration poses a significant challenge. In games where exploration is not as crucial we see that vanilla DQN often saturates the upper bound for game score. However, in these exploration allows for faster learning rates. 

\end{document}